\documentclass[10pt, a4paper]{article}

\usepackage[]{lrec-coling2024} 
\usepackage{multirow}
\usepackage{multicol}
\usepackage{booktabs}
\usepackage{amssymb}
\usepackage{xspace}

\newcommand{\ie}{\emph{i.e.,}\xspace}

\newcommand{\eg}{\emph{e.g.,}\xspace}

\newcommand{\etc}{\emph{etc}}
\newcommand{\ignore}[1]{}
\newcommand{\tabincell}[2]{\begin{tabular}{@{}#1@{}}#2\end{tabular}}

\title{ BAMBOO: A Comprehensive Benchmark for Evaluating Long Text Modeling Capacities of Large Language Models}

\name{Zican Dong$^{1}$, Tianyi Tang$^{1}$, Junyi Li$^{1,2}$, Wayne Xin Zhao$^{1}$\sthanks{~~Corresponding author.} , Ji-Rong Wen$^{1,3}$} 

\address{$^1$Gaoling School of Artificial Intelligence, Renmin University of China \\$^2$DIRO, Université de Montréal\\$^3$School of Information, Renmin University of China\\
         \{dongzican, steven\_tang, lijunyi, jrwen\}@ruc.edu.cn, batmanfly@gmail.com\\}

\abstract{
Large language models (LLMs) have achieved dramatic proficiency over NLP tasks with normal length. Recently, multiple studies have committed to extending the context length and enhancing the long text modeling capabilities of LLMs.
To comprehensively evaluate the long context ability of LLMs, we propose \textbf{BAMBOO}, a multi-task long context benchmark. BAMBOO has been designed with four principles: comprehensive capacity evaluation, avoidance of data contamination, accurate automatic evaluation, and different length levels. 
It consists of 10 datasets from 5 different long text understanding tasks, \ie question answering, hallucination detection, text sorting, language modeling, and code completion, to cover various domains and core capacities of LLMs.
We conduct experiments with five widely-used long-context models and further discuss five key questions for long text research.
In the end, we discuss problems of current long-context models and point out future directions for enhancing long text modeling capacities.
We release our data, prompts, and code at \url{https://github.com/RUCAIBox/BAMBOO}.
 \\ \newline  
}

\begin{document}

\maketitleabstract

\section{Introduction}


Recently, large language models (LLMs) have exhibited huge success across various NLP tasks~\cite{Ouyang-NIPS-2022-Training,Touvron-arxiv-2023-Llama,Zhao-2023-arxiv-A}. Despite their remarkable capacities, existing LLMs still face constraints due to their limited context length, suffering from performance decline when the input text exceeds their length limit. To enhance the capability of LLMs to model long texts, several studies extend the context window through modifying position embeddings~\cite{kaiokendev-github-2023-Things,Chen-arxiv-2023-Extending}, and continuing to fine-tune the models on multi-turn conversations or long text datasets~\cite{Zheng-arxiv-2023-Zheng}. Moreover, numerous research also explores transforming the original long text into short segments within the context length based on context compression techniques~\cite{Zhou-arxiv-2023-RecurrentGPT, jiang-arxiv-2023-longllmlingua}.

To push forward the research of long context modeling, it is crucial to establish an automatic, reliable, and comprehensive evaluation benchmark for assessing the performance of LLMs in tasks involving lengthy text.
In existing literature, several benchmarks have been proposed to evaluate the long context modeling ability of LLMs, such as ZeroSCROLLS~\citelanguageresource{Shaham-arxiv-2023-ZeroScrolls}, L-Eval~\citelanguageresource{An-arxiv-2023-L}, and LongBench~\citelanguageresource{Bai-arxiv-2023-LongBench}. However, a crucial aspect overlooked by these studies is the potential issue of data contamination~\cite{Golchin-arxiv-2023-Time}, \ie there might exist overlaps between the pre-training data and evaluation data. 
In the meanwhile, these benchmarks consist of some text generation tasks evaluated by automatic metrics such as BLEU~\cite{Wang-arxiv-2023-Is}. However, these metrics are blamed for their inaccurate evaluation of model performance.  


To this end, we propose \textbf{BAMBOO}, a comprehensive benchmark to analyze the long text modeling capacities of LLMs. 
In BAMBOO benchmark, we manually construct ten datasets from five tasks, including question answering, hallucination detection, text sorting, language modeling, and code completion. 
It comprehensively evaluates language generation, knowledge utilization, reasoning, and tool manipulation abilities over long texts. 
We aim to ensure that the salient information is spread throughout the complete long texts. 
Consequently, the LLM needs to model long-range dependencies to identify both the coarse-grained information related to the global comprehension of the complete text and fine-grained information about details across several sentences. 
Moreover, with datasets from multiple domains, BAMBOO can evaluate the performance of LLMs across various sources. 

\begin{table}[htb]

    \centering
    \resizebox{\columnwidth}{!}{

    \begin{tabular}{c|ccc|c}
    \toprule
        Benchmark&ZeroSCROLLS & L-Eval&LongBench&\textbf{BAMBOO}\\
        \midrule
        \tabincell{c}{Comprehensive Capacity\\Evaluation}&\checkmark&\checkmark&\checkmark&\textbf{\checkmark}
        \\\midrule
        \tabincell{c}{Accurate Automatic\\Evaluation}&$\times$&$\times$&$\times$&\textbf{\checkmark}
        \\\midrule
        \tabincell{c}{Avoidance of Data\\Contamination}&$\times$&$\times$&$\times$&\textbf{\checkmark}\\\midrule
        \tabincell{c}{Different\\Length Levels}&$\times$&$\times$&\checkmark&\textbf{\checkmark}
        \\\midrule
        \#Tasks&3&4&6&\textbf{5}
        \\
         \bottomrule
    \end{tabular}}
    \caption{Comparison with other long text evaluation benchmarks. \#Tasks denotes the number of tasks in the benchmark.}
    \label{table_compare}
\end{table}

BAMBOO is divided into two subsets at \textit{different length levels}, \ie BAMBOO-4k and BAMBOO-16k, according to the number of tokens in the prompt\footnote{All references to the number of tokens in our paper are computed using the gpt-3.5-turbo tokenizer.}. Each subset has 1502 samples with an average length of 3152 and 7500 tokens, respectively. To alleviate the \textit{data contamination} with pre-training corpora, BAMBOO is constructed based on the data sources released in 2023.
Furthermore, we select diverse tasks with \textit{accurate automatic metrics} for reliable evaluation. 
Table~\ref{table_compare} presents a comparison of BAMBOO with existing benchmarks.

We extensively evaluate five long-context LLMs on the BAMBOO benchmark. We observe that ChatGPT-16k consistently demonstrates optimal performance across most datasets, whereas other models usually struggle, especially on uncommon tasks. 
We conduct on a series of analytical experiments to shed light on the key questions in long text modeling. Our findings are listed as follows:

$\bullet$ Extending the context window of LLMs is a double-edged sword, which is beneficial for medium-length texts but harmful for short texts.

$\bullet$ Using evidence and properly compressing long texts can generally enhance the performances of LLMs. However, smaller LLMs still exhibit limited performance even given more concise input.

$\bullet$ LLMs' performances on different datasets are sensitive to positions of instructions. Additionally, locating evidence on both ends is beneficial due to larger attention scores.



Ultimately, we undertake a discussion of problems associated with instruction forgetting, format errors, reasoning capabilities, and uncommon tasks on long text modeling. We conclude by identifying future research directions for advancing the long text modeling capabilities of LLMs.

\section{Related Work}

\paragraph{Long-Context Architecures}

Owing to the quadratic memory and computational complexity of self-attention, original Transformer models encountered significant challenges in scaling to extended contexts. To enhance the efficiency of Transformers, researchers explored multiple innovations in the attention mechanism. These include recurrent Transformers~\cite{Dai-ACL-2019-Transformer,Rae-ICLR-2020-Compressive}, local attention~\cite{Child-arxiv-2019-generating,Baltagy-arxiv-2020-Longformer,Zaheer-NAACL-2020-Big}, content-based sparse attention~\cite{Kitaev-ICLR-2020-Reformer,Tay-ICML-2020-Sparse,Roy-Trans-2021-Efficient}, and attention approximation~\cite{Wang-arxiv-2020-Linformer,Ma-NIPS-2021-Luna, Katharopoulos-ICML-2020-Transformers,Peng-ICLR-2021-Random}.
In addition to the above Transformer variants, several models were introduced, replacing attention with alternative modules. S4~\cite{Gu-ICLR-2022-Efficiently}, DSS~\cite{Gupta-NIPS-2022-Diagonal}, and GSS~\cite{Mehta-ICLR-2023-Long} employed state space models, while RetNet~\cite{Sun-arxiv-2022-Retentive} made use of retention mechanism, achieving parallel training and recurrent inference for long sequences.


\begin{table*}[htb]
    \centering

    \resizebox{2 \columnwidth}{!}{
    \begin{tabular}{lccccccc}
    \toprule
    \textbf{Dataset} & \textbf{\#Input}&\textbf{\#Example}&\textbf{Metric} & \textbf{Domain} & \textbf{Source}
    \\\midrule
    AltQA&3243/13084&200/200&accuracy&Wikipedia&\url{https://github.com/abacusai/long-context}
    \\
    PaperQA&3101/6838&100/100&accuracy&Paper&\url{https://aclanthology.org/}\\
    MeetingQA&2738/9838&100/100&accuracy&Meeting&\url{https://record.assembly.wales/}\\
    \midrule
        SenHallu &3170/6357&200/200&P/R/F1&Paper &\url{https://aclanthology.org/} \\
         AbsHallu&3314/6445&200/200&P/R/F1&Paper&\url{https://aclanthology.org/}

         \\\midrule
        ShowsSort&2992/6411&200/200&CI&TV Shows&\url{https://tvmeg.com/}\\
        ReportSumSort&3753/8309&150/150&CI&Reports&\url{https://www.gao.gov/}\\

        \midrule
        ShowsPred&2389/4860&100/100&accuracy&TV Shows&\url{https://tvmeg.com/}\\
        MeetingPred&3689/11578&100/100&accuracy&Meeting&\url{https://record.assembly.wales/}\\
        \midrule
        PrivateEval&3149/6230&152/152&pass@1&Code&\url{https://github.com/microsoft/PyCodeGPT}

         \\\bottomrule
    \end{tabular}}
    
    \caption{Overview of our BAMBOO benchmark. \#Input and \#Example represent the average number of tokens in the input text and the number of examples of each dataset, respectively. For \#Input and \#Example,  ``number1/number2'' is the number of the middle and long levels, respectively. P/R/F1 denotes precision/recall/F1, and CI stands for concordance index.}
    \label{table_overview}
\end{table*}
\paragraph{Adapting LLMs for Long Contexts}
Recently, with the booming of LLMs, there has been an increase in research efforts for overcoming the length limitations of off-the-shelf LLMs. 
Through modifying RoPE~\cite{Su-arxiv-2021-RoFormer} with position interpolation~\cite{kaiokendev-github-2023-Things, Chen-arxiv-2023-Extending}, truncated basis~\cite{Pal-arxiv-2023-Giraffe}, or modifying bases~\cite{xiong-arxiv-2023-effective, peng-arxiv-2023-yarn,Liu-arxiv-2023-Scaling,baptiste-arxiv-2023-code}, the context windows of LLMs could be extended to 32k or more tokens. 
Equipped with fusion-in-decoder~\cite{Gautier-EACL-2020-Leveraging,Ratner-ACL-2023-Parallel} or external memory techniques~\cite{Bertsch-arxiv-2023-Unlimiformer,Tworkowski-arxiv-2023-Focused,Wang-arxiv-2023-Augmenting}, LLMs could access tokens from a distant past during the generation stage. 
In addition, context compression techniques, such as retrieval-augmentation and recurrence, could significantly expand their capacity to model longer contexts, where LLMs only need to process a short segment at once~\cite{Zhou-arxiv-2023-RecurrentGPT,liang-arxiv-2023-unleashing,packer-arixv-2023-memgpt,xu-arxiv-2023-retrieval,jiang-arxiv-2023-longllmlingua}.

\paragraph{Evaluation for Long Text Modeling}

For evaluating long text modeling capabilities, diverse datasets were proposed, mainly classified into three task categories: language modeling, question answering, and summarization. 
The objective of language modeling was to sequentially predict the next tokens based on previous context~\cite{Rae-ICLR-2020-Compressive}. However, popular benchmarks did not adopt it since most tokens could be easily predicted within a limited context~\cite{Sun-EMNLP-2021-Do}. 
Long text summarization datasets encompassed normal summarization~\citelanguageresource{Cohan-NAACL-2018-A,Huang-NAACL-2021-Efficient,Chen-ACL-2022-SummScreen} and query-based summarization~\citelanguageresource{zhong-ACL-2021-QMSUM}, involving the condensation of either the entire or a portion of lengthy texts into concise summaries.
As for question answering, given a question and a single document~\citelanguageresource{Pang-NAACL-2022-QUALITY,Dasigi-acl-2021-a} or multiple documents~\citelanguageresource{yang-EMNLP-2018-HotpotQA}, the model needed to provide the answer of the question.


To facilitate comprehensive evaluations of long text modeling abilities, several long text benchmarks were introduced. 
Scrolls~\citelanguageresource{Shaham-ACL-2022-Scrolls} collected and standardized several long text reasoning datasets spanning various tasks and domains into a unified format. 
In the wake of the heightened interest in LLMs, ZeroScrolls~\citelanguageresource{Shaham-arxiv-2023-ZeroScrolls}, L-Eval~\citelanguageresource{An-arxiv-2023-L}, and LongBench~\citelanguageresource{Bai-arxiv-2023-LongBench} were released to assess the zero-shot long text modeling ability of LLMs. Nevertheless, they all encountered potential data contamination problems and suffered from inaccurate automatic metrics.

%

\section{BAMBOO Benchmark}
\label{section_benchmark}

Our BAMBOO benchmark commits to equitably and thoroughly evaluating LLMs' long text modeling proficiency. 
It contains ten elaborately constructed datasets across a broad range of domains and evaluations. The overview of the BAMBOO benchmark is shown in Table~\ref{table_overview}.

\subsection{Design Principles}

Different from existing long text benchmarks in Table~\ref{table_compare}, our BAMBOO benchmark is meticulously designed to comprehensively evaluate the long text modeling capacity of LLMs. Our benchmark can potentially circumvent the issue of data contamination in previous benchmarks, pursue more accurate automatic evaluation, and cater to LLMs at different length levels.

\paragraph{Comprehensive Capacity Evaluation}

The primary objective of BAMBOO is comprehensively evaluating the abilities of LLMs in modeling long texts via different tasks and domains. To meet the real needs in long text scenarios, all the tasks assess the LLMs' ability to capture long-range dependencies. Inspired by \citet{Zhao-2023-arxiv-A}, BAMBOO covers diverse tasks evaluating language generation, knowledge utilization, reasoning, and tool manipulation capacities over long texts. BAMBOO can also evaluate the coarse-grained comprehension of entire texts and the fine-grained reasoning concerning specific details. Moreover, we collect data from various sources, assessing and comparing abilities across diverse domains.

\paragraph{Avoidance of Data Contamination}
In BAMBOO, we make a deliberate effort to minimize the overlap between test data and the training corpora of LLMs as much as possible. 
One of the key challenges when evaluating the long text capability of LLMs is the model has ``seen'' the input text during the training stages, \ie data contamination~\cite{Golchin-arxiv-2023-Time}. 
To alleviate this issue, we take a cautious approach by retaining data only released in 2023, since training data cutoff for gpt-3.5-turbo and Claude2 are up to September 2021\footnote{\url{https://platform.openai.com/docs/models/gpt-3-5}\label{footnote_chatgpt}} and the early 2023\footnote{\url{https://www.anthropic.com/index/claude-2}\label{claude}}, respectively. 
Yet, since the specifics of their training data remain undisclosed for some models, we cannot guarantee the absence of any pre-training data used in our benchmark.
In addition, we also utilize datasets with old data sources following~\citet{Pal-arxiv-2023-Giraffe}, but any appearances of the answers have been modified. Thus, the model must respond based solely on the input, reducing the potential effect of exposure to similar data.

\paragraph{Accurate Automatic Evaluation}
Each task in BAMBOO is designed for precise automatic evaluation.
Previous benchmarks include text generation tasks (\eg text summarization and question answering). The generations of LLMs may exhibit variations in expression compared to the references, making it challenging to accurately evaluate their outputs with automatic metrics~\cite{Tang-arxiv-2023-not}, even using LLMs as evaluator~\cite{Wang-arxiv-2023-Is}. 
Hence, we transform some generation tasks into multi-choice tasks
and select tasks that can be precisely evaluated. All the tasks can be measured by metrics such as accuracy, pass@1, \etc.

\paragraph{Different Length Levels}
In BAMBOO, each task encompasses two distinct length levels. 
The most popular LLMs, \eg Llama2~\cite{Touvron-arxiv-2023-Llama2} and gpt-3.5-turbo\textsuperscript{\ref{footnote_chatgpt}}, only have a context window of 4k tokens while some of them extend their context windows to 16k or more tokens, including gpt-3.5-turbo-16k\textsuperscript{\ref{footnote_chatgpt}} and ChatGLM2-6b-32k~\cite{Zeng-ICLR-2023-GLM}.
However, previous benchmarks did not thoroughly consider different length distributions of different datasets, making it difficult to analyze the impact of length changes on LLMs in the same task.
Instead, we divide each task of BAMBOO into two subsets based on the input length, with the maximum length of 4k and 16k tokens, respectively. Thus, we can effectively measure the impact of lengths on the proficiency of LLMs.


\subsection{Data Collection}

To ensure domain diversity, we collect long texts from four distinct sources, namely NLP research papers, government reports, TV show transcripts, and committee meetings transcripts. 
All these data sources are publicly accessible and released in 2023 to avoid pre-training data contamination. We crawl and parse the source websites, extracting only the plain text while excluding images, tables, and footnotes. Texts containing fewer than 1000 tokens are excluded, and overly lengthy texts are truncated. 
In addition to these latest corpora, we also consider the pseudo long-context datasets where potential data contamination has attempted to be avoided. 



In pursuit of the objective of encompassing diverse length levels, we divide samples within the BAMBOO benchmark into two subsets based on the length of the complete input text. One subset contains prompts up to 4k tokens (BAMBOO-4k), while the other comprises prompts with tokens from 4k to 16k (BAMBOO-16k). 

\subsection{Task Construction}
We aim to assess long text modeling ability in BAMBOO thoroughly. To attain the goal, we devise five distinct tasks and comprise ten corresponding datasets based on the fundamental principle. 

\paragraph{Question Answering}
In the question answering task, we primarily focus on evaluating knowledge utilization and reasoning ability. 
With the assistance of human labelers, we manually construct two multi-choice question answering datasets, \textbf{PaperQA} and \textbf{MeetingQA}, to evaluate the LLMs' comprehension of a long document or dialogue. We ask annotators to rephrase the expressions of questions and options instead of directly copying the original expressions in given texts. We also ensure that most of the answers need to be reasoned from evidence from multiple paragraphs.
In addition, we utilize \textbf{AltQA}~\cite{Pal-arxiv-2023-Giraffe}, comprising multiple Wikipedia documents. Each question requires a numerical answer, and every occurrence of the answer within the documents is modified to another number to avoid data contamination. 




\paragraph{Hallucination Detection} 
Hallucination is a prevalent phenomenon in LLMs, \ie the generated content may conflict with or be beyond the source. In this task, models must utilize the knowledge of given contexts to detect hallucinations. We create two novel datasets, \ie \textbf{SenHallu} and \textbf{AbsHallu}, using ChatGPT to generate hallucinations based on the correct hypothesis (\eg a few words modification or a make-up sentence insertion)~\cite{Li-arxiv-2023-HaluEval}. Based on the concept of natural language inference tasks, we present LLMs with a paper and a hypothesis. Then, we request LLMs to determine whether the hypothesis entails or contradicts the paper as a binary classification task. 



\paragraph{Text Sorting}

To evaluate model's capacity of reasoning over the logical order of texts, we devise two text sorting datasets that require LLMs to reorder the shuffled texts according to contextual information of the whole texts. 
\textbf{ShowsSort} is a sorting task involving shuffled segments of a TV show transcript. The original order is reconstructed based on contextual continuity and the inherent meaning of each segment.
\textbf{ReportSumSort} is another sorting task with the input of a complete government report and a list of shuffled summaries. The objective is to reorder these summaries according to their positions within the segments represented in the report.




\paragraph{Language Modeling} 
Language modeling task is a typical language generation task where the next token is sequentially predicted based on the past context. 
In order to evaluate the capacity to handle long-range dependencies, we design the task intending to predict the last utterance's speaker within a long dialogue.
Specifically, we adopt multi-turn conversation as the source and format each utterance as ``\{utterance\} said by \{speaker\}''. 
Then, we remove the speaker of the last turn of conversation as the target. Successfully predicting the speaker necessitates the model's grasp of the characteristics of existing speakers and the contextual conversations. We construct \textbf{ShowsPred} and \textbf{MeetingPred} with transcripts of TV shows and meetings as sources, respectively. 

\begin{table*}[htb]
\renewcommand\arraystretch{1.1}
\setlength\tabcolsep{2.5pt}
    \centering
     
\resizebox{1.9\columnwidth}{!}{
\begin{tabular}{lcccccccccc}
\toprule
\multirow{2.5}{*}{\textbf{Models}}   & \multicolumn{2}{c}{\textbf{ShowsSort}} & \multicolumn{2}{c}{\textbf{ReportSumSort}}   & \multicolumn{2}{c}{\textbf{ShowsPred}} & \multicolumn{2}{c}{\textbf{MeetingPred}}& \multicolumn{2}{c}{\textbf{SenHallu}} \\ 
\cmidrule(r){2-3}\cmidrule(r){4-5}\cmidrule(r){6-7}
\cmidrule(r){8-9}\cmidrule(r){10-11}
& 16k & 4k & 16k & 4k & 16k & 4k & 16k & 4k & 16k & 4k \\
\midrule
gpt-3.5-turbo-16k&\textbf{59.0}&\textbf{58.2}&\textbf{69.6}&\textbf{77.6}&\textbf{55.0}&\textbf{53.0}&\textbf{57.0}&\textbf{73.0}&62.7/99.0/76.7&68.1/96.0/79.7\\
Claude2-100k&53.1&54.0&59.7&66.7&8.0&26.0&0.0&21.0&\textbf{67.3}/99.0/\textbf{80.2}&\textbf{69.3}/97.0/\textbf{80.8}\\
\midrule
ChatGLM2-32k&32.0&42.7&31.6&43.5&13.0&8.0&14.0&25.0&52.9/\textbf{100.0}/69.2&52.7/96.0/68.1\\
Vicuna-v1.5-16k&53.2&53.6&50.1&51.4&5.0&11.0&5.2&64.0&56.4/92.9/70.2&67.7/99.0/69.3\\
Longchat-v1.5-16k&53.2&53.8&44.5&47.9&1.0&1.0&29.9&36.0&50.3/\textbf{100.0}/66.9&51.2/\textbf{100.0}/67.8\\
\midrule
\textcolor{gray}{Random Baseline}&\textcolor{gray}{50.0}&\textcolor{gray}{50.0}&\textcolor{gray}{50.0}&\textcolor{gray}{50.0}&\textcolor{gray}{7.6}&\textcolor{gray}{10.3}&\textcolor{gray}{8.0}&\textcolor{gray}{13.9}&\textcolor{gray}{50.0/50.0/50.0}&\textcolor{gray}{50.0/50.0/50.0}
\\

\midrule[0.8pt]
\multirow{2.5}{*}{\textbf{Models}}&\multicolumn{2}{c}{\textbf{PrivateEval}}&\multicolumn{2}{c}{\textbf{AltQA}} & \multicolumn{2}{c}{\textbf{PaperQA}}&\multicolumn{2}{c}{\textbf{MeetingQA}}& \multicolumn{2}{c}{\textbf{AbsHallu}}\\ 
\cmidrule(r){2-3}\cmidrule(r){4-5}\cmidrule(r){6-7}
\cmidrule(r){8-9}\cmidrule(r){10-11}
& 16k & 4k & 16k & 4k & 16k & 4k & 16k & 4k & 16k & 4k \\
\midrule

gpt-3.5-turbo-16k&\textbf{21.7}&\textbf{21.1}&\textbf{72.0}&\textbf{76.5}&\textbf{75.0}&\textbf{78.0}&\textbf{72.0}&\textbf{75.0}&55.6/\textbf{100.0}/71.4&57.2/99.0/72.5 \\
Claude2-100k&0.7&7.2&4.5&27.0&51.0&69.0&47.0&63.0&\textbf{56.8}/\textbf{100.0}/\textbf{72.5}&\textbf{58.2}/99.0/\textbf{73.3}\\
\midrule
ChatGLM2-32k&3.5&0.7&67.0&64.0&65.0&62.0&65.0&52.0&50.5/\textbf{100.0}/66.7&50.0/99.0/66.4\\
Vicuna-7b-v1.5-16k&2.0&3.3&25.0&30.5&37.0&29.0&27.0&31.0&51.0/99.0/67.3&53.3/88.0/66.4\\
Longchat-7b-v1.5-16k&0.7&3.9&41.7&64.0&44.0&44.0&36.0&42.0&50.3/\textbf{100.0}/66.9&50.0/\textbf{100.0}/66.7\\
\midrule
\textcolor{gray}{Random Baseline}&\textcolor{gray}{0.0}&\textcolor{gray}{0.0}&\textcolor{gray}{0.0}&\textcolor{gray}{0.0}&\textcolor{gray}{25.0}&\textcolor{gray}{25.0}&\textcolor{gray}{25.0}&\textcolor{gray}{25.0}&\textcolor{gray}{50.0/50.0/50.0}&\textcolor{gray}{50.0/50.0/50.0}
\\

\bottomrule
\end{tabular}
}
\caption{Results (\%) of selected long-context LLMs on our benchmark. We report both the performance on 4k and 16k versions of all datasets. For SenHallu and AbsHallu, we sequentially exhibit precision, recall, and F1.
}
\label{table_main_results}
\end{table*}

\paragraph{Code Completion}
To assess the capability of LLMs to effectively utilize external tools through API calls in tackling complex tasks, we construct the \textbf{PrivateEval} dataset on the basis of the benchmark from~\citetlanguageresource{Zan-EMNLP-2022-When}. In the dataset, LLMs are presented with API documents derived from private libraries and a partial code snippet. Then, the model needs to identify key API documents to complete the code snippet.
To build the dataset, we manually alter the keywords within API documents of the Torchdata library following previous methods~\citelanguageresource{Zan-EMNLP-2022-When} on Monkey and BeatNum libraries. Furthermore, we adjust the number of the provided documents for length control. 

\subsection{Metrics}
As shown in Table~\ref{table_overview}, we employ four metrics to evaluate different tasks in our benchmark, \ie accuracy for question answering and language modeling, concordance index~\citelanguageresource{Shaham-arxiv-2023-ZeroScrolls} for sorting, pass@1~\cite{Chen-arxiv-2021-Evaluating} for code completion, and precision, recall, and F1 for hallucination detection. Notably, all the answers are generated freely, and any answer with a wrong format is treated as incorrect.





\subsection{Benchmark Usage}

The BAMBOO benchmark offers researchers a comprehensive platform to thoroughly evaluate the long text capabilities of LLMs. 
First, with multiple tasks covering the evaluation of different capacities, BAMBOO provides a direct comparison of different long-context LLMs' comprehensive capacities. 
Meanwhile, the BAMBOO benchmark enables a detailed analysis of LLMs' weaknesses in different aspects. Thus, more powerful long-context LLMs can be built by making up for these deficiencies. 
Finally, the BAMBOO benchmark introduces varying length levels, enabling assessing the significance of the lengths of prompts and context windows.
In total, BAMBOO can promote better handling of the multifaceted challenges and opportunities in the realm of long text modeling.

Users can utilize the BAMBOO benchmark with instructions provided in our project repository or instructions for your design. Then, users can evaluate the generations with our code.

\section{Experiment}
\label{sec_experiment}

With the BAMBOO benchmark, we conduct experiments to evaluate the zero-shot long text modeling capability of long-context LLMs. Additionally, we further delve into key questions of long text modeling and discuss problems of long-context models.

\subsection{Baselines}

We select five instruction-tuned models with a context length of over 16k tokens as our baselines. For closed models available via API, we select gpt-3.5-turbo-16k\textsuperscript{\ref{footnote_chatgpt}} from OpenAI and Claude2-100K\textsuperscript{\ref{claude}} from Anthropic. For open-sourced models, we select Vicuna-7b-v1.5-16k~\cite{Zheng-arxiv-2023-Zheng}, Longchat-7b-v1.5-32k~\cite{li-github-2023-longchat}, and ChatGLM2-6b-32k~\cite{Du-ACL-2022-GLM,Zeng-ICLR-2023-GLM}~\footnote{We abbreviate gpt-3.5-turbo as ChatGPT, Vicuna-7b-v1.5 as Vicuna, Claude2-100k as Claude2, Longchat-7b-v1.5 as Longchat, and ChatGLM2-6b as ChatGLM2 in the following.}.

We add a random baseline for each task. For text sorting and language modeling tasks, we randomly return the permutated sequences and pre-existing speakers in the dialogues, respectively. For hallucination detection and multi-choice QA datasets, we randomly select one choice from (true, false) or (A, B, C, D).
For PrivateEval and AltQA, we directly set the random baseline to 0 due to the infinite search space for answers. 








\subsection{Overall Results}

The overall results for the model performance on the BAMBOO benchmark are displayed in Table~\ref{table_main_results}. Across different datasets and length levels, ChatGPT-16k consistently exhibits excellent performance, surpassing other LLMs in nearly all tasks, except a minor performance gap observed in the hallucination detection task. Conversely, the other models' performances are often poor, even falling below performances of random baselines.


Moreover, models usually struggle with datasets with uncommon objectives and complex requirements, even the most powerful ChatGPT. In hallucination detection tasks where distinctions between the hallucinated and correct hypotheses are subtle, most LLMs can hardly capture those hallucinations, resulting in low precision. In addition, LLMs also underperform in ShowsSort and PrivateEval, highlighting their inability to aggregate, manipulate tools, and synthesize code.
We infer that limited diversity of training data types plays a role in the subpar performance of LLMs.

Comparing performances across different length levels, a notable trend is that there is a discernible decrease in performance for most datasets and models as extending input length. Due to the incorrect generating format, the trend is more prominent for tasks with diverse outputs.

\subsection{Key Questions of Long Text Modeling}

\label{sec_analysis}
To delve into the mechanism and influential factors affecting the long text modeling capacities of LLMs, our investigation centers on five key research questions. In this section, we conduct analytical experiments to address these questions\footnote{We only report F1 score for hallucination detection tasks followingly.}.



\subsubsection*{RQ1: Do LLMs with Long Context Windows Pay for Extension Tax?}

We first discuss whether extending the LLMs' context windows and fine-tuning LLMs on long texts harm performance over tasks with shorter texts, \ie ``extension tax''. We compare ChatGPT-16k and Vicuna-16k to their short-context variants on selected tasks from BANBOO-4k and MMLU~\citelanguageresource{Hendrycks-ICLR_2021-MMLU}.

\begin{figure}[htb]
    \centering
    \resizebox{\columnwidth}{!}{
    \includegraphics{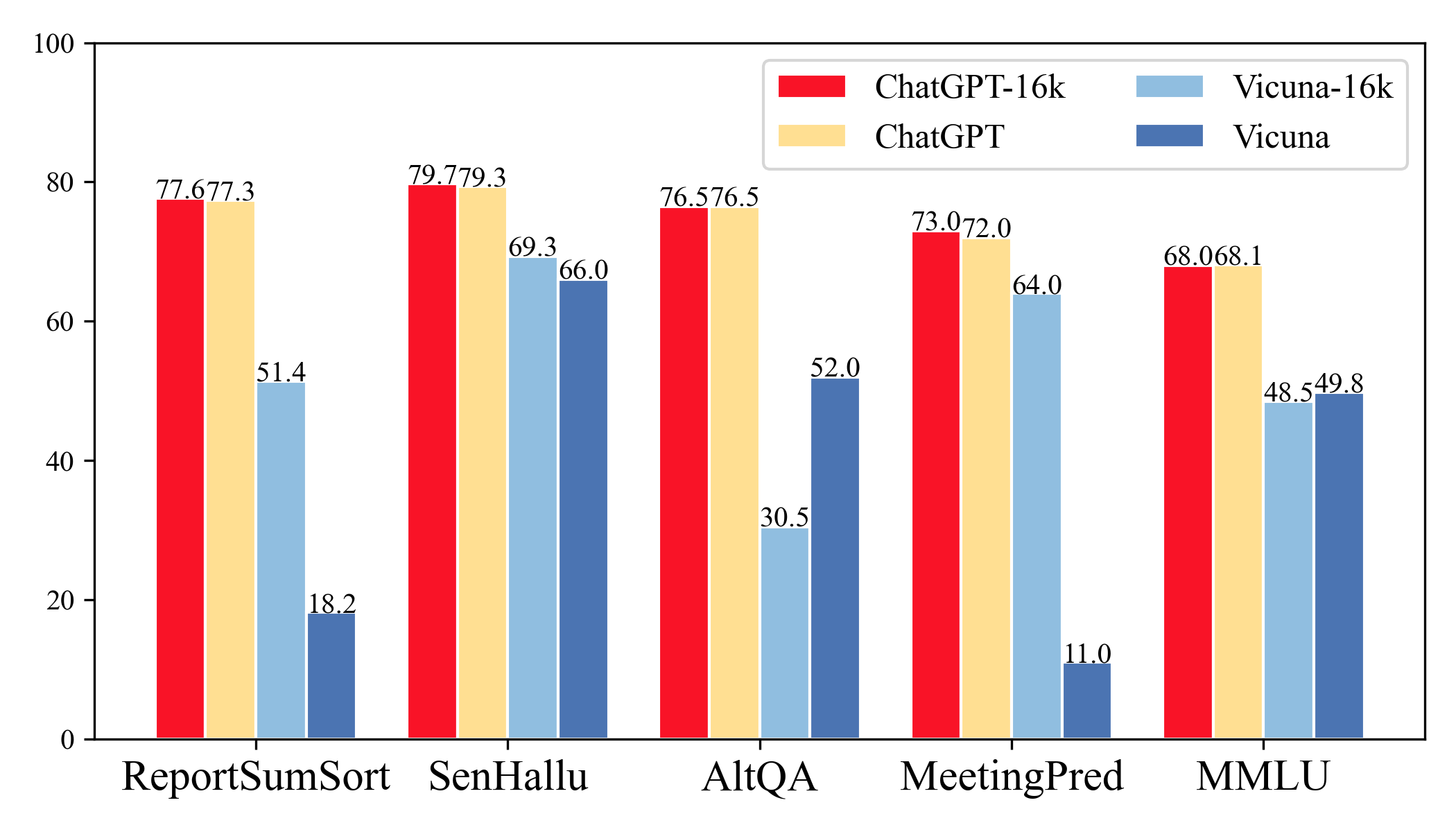}}
    \caption{Results of long and short-context LLMs on BAMBOO-4k and MMLU.}
    \label{fig_experiment_length}
\end{figure}

As shown in Figure~\ref{fig_experiment_length}, ChatGPT and Vicuna exhibit disparate trends. Performances between ChatGPT and ChatGPT-16k over different tasks are almost identical. However, an ``extension tax" is evident when replacing Vicuna with Vicuna-16k on MMLU. The longer training data and position interpolation detrimentally affect the ability of short-context tasks. Interestingly, Vicuna-16k demonstrates improved performance in BAMBOO-4k, suggesting that longer training data can enhance performance in medium-length tasks.

\subsubsection*{RQ2: To What Extent Do LLMs Struggle Due to The Long Input?}
In the long text scene, a critical question emerges: is the challenge associated with the length of the text or the inherent capacity to solve the task? In MeetingQA, PaperQA, and PrivateEval datasets, only concise evidence in the long text is sufficient to support the completion of tasks. Within these datasets, we make a comparison of the performance between inputting only evidence and the complete text.




\begin{table}[htb]
    
    \resizebox{\columnwidth}{!}{
    \begin{tabular}{lccccccc}
    \toprule
         \multirow{2.5}{*}{\textbf{Models}}   & 
         \multirow{2.5}{*}{\textbf{Input}}&\multicolumn{2}{c}{\textbf{MeetingQA}} & \multicolumn{2}{c}{\textbf{PaperQA}}   & \multicolumn{2}{c}{\textbf{PrivateEval}}  \\ 
\cmidrule(r){3-4}\cmidrule(r){5-6}\cmidrule(r){7-8}  
&&16k & 4k&16k & 4k&16k & 4k\\\midrule
         \multirow{2}{*}{ChatGPT-16k}&evidence&\textbf{74.0}&\textbf{78.0}&\textbf{76.0}&\textbf{79.0}&\textbf{31.6}&\textbf{31.6}\\
         &complete&72.0&75.0&75.0&78.0&21.7&21.1\\\midrule
         \multirow{2}{*}{Vicuna-16k}&evidence&\textbf{40.0}&\textbf{36.0}&35.0&\textbf{36.0}&\textbf{5.3}&\textbf{5.3}\\
         &complete&27.0&31.0&\textbf{37.0}&29.0&2.0&3.3
         \\\bottomrule
         
    \end{tabular}}
    \caption{Comparison of LLMs' performances with the input of only evidence or the complete text. }
    \label{table_golden}
\end{table}

\begin{table*}[htb]
    \centering
    \resizebox{2\columnwidth}{!}{

\begin{tabular}{llcccccccccccccc}
    \toprule       
    \multirow{2.5}{*}{\textbf{Model}}   & \multirow{2.5}{*}{\textbf{Position}} & \multicolumn{2}{c}{\textbf{ShowsSort}}& \multicolumn{2}{c}{\textbf{SenHallu}}&\multicolumn{2}{c}{\textbf{AbsHallu}} & \multicolumn{2}{c}{\textbf{MeeetingPred}}& \multicolumn{2}{c}{\textbf{ShowsPred}}& \multicolumn{2}{c}{\textbf{MeeetingQA}}& \multicolumn{2}{c}{\textbf{PaperQA}}\\
    \cmidrule(r){3-4}\cmidrule(r){5-6}
\cmidrule(r){7-8}\cmidrule(r){9-10}\cmidrule(r){11-12}\cmidrule(r){13-14}\cmidrule(r){15-16}
    && 16k & 4k & 16k & 4k & 16k & 4k& 16k & 4k& 16k & 4k& 16k & 4k& 16k & 4k
    \\\midrule
         \multirow{3}{*}{ChatGPT-16k}&Pre-Ins&54.5&56.4&\textbf{76.7}& \textbf{79.7}&69.3&70.3&\textbf{20.5}&\textbf{45.0}&\textbf{35.0}&\textbf{33.0}&78.0&73.0&70.0&71.0\\
         &Post-Ins&\textbf{55.4}&54.6&75.8&77.3&\textbf{69.7}&\textbf{71.9}&9.5&16.0&14.0&20.0&75.0&\textbf{80.0}&\textbf{74.0}&77.0\\
         &Both-Ins&54.6&\textbf{60.2}&75.8&75.5 &68.7&70.7&16.0& 29.0&26.0&31.0&\textbf{81.0}&79.0&73.0&\textbf{80.0}\\
         \midrule
         \multirow{3}{*}{Vicuna-16K}&Pre-Ins&42.8&\textbf{53.9}&66.7&70.5&66.7&66.4&5.0&\textbf{24.0}&\textbf{6.0}&\textbf{16.0}&6.0&\textbf{32.0}&\textbf{33.0}&24.0\\
         &Post-Ins&51.4&51.4&67.8&\textbf{73.7}&\textbf{66.9}&66.7&\textbf{13.0}&2.0&0.0&2.0&21.0&27.0&24.0&\textbf{38.0}\\
         &Both-Ins&\textbf{52.4}& 53.4&\textbf{67.9}  &66.7&66.7&\textbf{69.0}& 12.0&8.0&0.0&4.0&\textbf{24.0}&29.0&32.0&29.0
         \\
         \midrule
        \multirow{3}{*}{LongChat-32k}&Pre-Ins& \textbf{51.5}&\textbf{53.8}&\textbf{66.7}&67.6&66.2&66.7&12.0&\textbf{37.0}&\textbf{4.0}&\textbf{21.0}&14.0&\textbf{37.0}&28.0&29.0\\
        &Post-Ins &49.8&48.9 &\textbf{66.7}&66.9&66.7&66.7&\textbf{19.0}&19.0&1.0&8.0&\textbf{34.0}&33.0&\textbf{30.0}&\textbf{39.0}\\
        & Both-Ins &50.3&\textbf{53.8}&\textbf{66.7}&\textbf{68.7}&\textbf{67.1}&\textbf{67.6}&18.0&18.0&4.0&8.0&32.0&34.0&24.0&32.0 \\
        \midrule

        \multirow{3}{*}{ChatGLM2-32k}&Pre-Ins&\textbf{42.6}&45.4&69.2&69.5&66.4&69.4&\textbf{13.0}&\textbf{18.0}&\textbf{9.0}&\textbf{15.0}&34.0&46.0&47.0&54.0\\
         &Post-Ins&29.2&45.7&\textbf{69.7}&\textbf{71.2}&\textbf{67.3}&\textbf{70.2}&8.0&11.0&7.0&\textbf{15.0}&\textbf{61.0}&\textbf{63.0}&\textbf{65.0}&66.0\\
         &Both-Ins&37.8&\textbf{47.5}&69.4&67.8 &66.9&69.0&12.0&15.0&\textbf{9.0}&11.0&54.0&56.0&61.0&\textbf{67.0}\\
         \bottomrule
\end{tabular}}

\caption{Results of LLMs with three different instruction positions relative to content. 
}
\label{table_position}
\end{table*}
Table~\ref{table_golden} illustrates a general trend of improved performance across various datasets. Mislocating evidence within long text contributes to some errors. Nevertheless, the improvements are relatively minor, and Vicuna's performance even declines in PaperQA. Additionally, the model consistently commits the same errors in both settings. Therefore, we infer that the awful performances of LLMs are predominantly attributed to subpar reasoning and coding ability rather than the mislocalization of evidence in the long text.



\subsubsection*{RQ3: How Do Instruction Positions Affect Long Text Modeling?}

A recent study observed that positioning the instruction at the end of the input can enhance generation performance~\cite{Liu-arxiv-2023-Instruction}.
Therefore, we explore the influence of instruction placement in long text tasks. Specifically, we segment the complete prompt into two distinct components:
``context'' encompasses the long text, and ``instruction'' contains task descriptions, optional questions, hypotheses, \etc. Following~\citet{Liu-arxiv-2023-Instruction}, We respectively position the instruction at the beginning, end, and both ends of the prompt, denoted as ``Pre-Ins," ``Post-Ins," and ``Both-Ins." We present a comparison of their effects in Table~\ref{table_position}.

To start with, when the instruction is exclusively placed at the beginning, a discernible decline in performance is commonly encountered as extending input lengths. Notably, the ``Pre-Ins" configuration within BAMBOO-16k often exhibits problems related to forgetting instructions, which is potentially attributed to the long-range decay of attention scores of RoPE~\cite{xiong-arxiv-2023-effective}.



Moreover, the optimal instruction positions vary depending on the datasets and models. ``Pre-Ins" typically result in lower performance for tasks where the instructions contain content-relevant information, such as question answering and hallucination detection. Conversely, ``Post-Ins" exhibit reversed trends, consistent with findings from previous research~\cite{Liu-arxiv-2023-Instruction}. Furthermore, ``Both-Ins" consistently yield sub-optimal results across most datasets, with only minor performance gaps compared to the best-performing strategies.





\subsubsection*{RQ4: Can LLMs Model Longer Text with Context Compression Methods?}


Prior research has effectively leveraged context compression techniques such as retrieval and truncation to address long text tasks with pretrained language models~\cite{Gong-ACL-2020-Recurrent,Zhang-ACL-2022-SUMMN,Zhao-arxiv-2022-Dense}. Thus, it holds promise to integrate these techniques with LLMs.

To verify whether context compression techniques can enhance long text modeling abilities of short-context LLMs, we employ three different methods for hallucination detection and question answering tasks. (1) \textit{Truncation}: We truncate the initial 3000 tokens. (2) \textit{Retrieval}: we partition the long text into segments with 256 tokens, representing the question and each segment into an embedding with openai's text-embedding-ada-002\footnote{\url{https://platform.openai.com/docs/guides/embeddings}}. Then, we retrieve top-10 relevant segments to the question. (3) \textit{Summarization}: We partition the input into chunks with 1024 tokens, sequentially summarize each chunk, and concatenate these summaries to form the final input.
\begin{table}[htb]
    \centering
    \resizebox{\columnwidth}{!}{
    \begin{tabular}{lccccc}
    \toprule
        \textbf{Method} &\multicolumn{1}{c}{\textbf{AQA}} &\multicolumn{1}{c}{\textbf{PQA}} & \multicolumn{1}{c}{\textbf{MQA}}   & \multicolumn{1}{c}{\textbf{SHallu}}&\multicolumn{1}{c}{\textbf{AHallu}}
         \\\midrule
         ChatGPT-16K&\textbf{72.0}&75.0&72.0&  76.7&\textbf{71.4} \\
        ChatGPT+Retrieval& 69.5 & \textbf{79.0} &\textbf{78.0}&\textbf{78.3}&71.2 \\
         ChatGPT+Truncation&71.5  & 64.0 &68.0& 74.1 &70.0  \\
        ChatGPT+Summarization&  51.5&63.0  &67.0  &75.0  &70.0\\
         \midrule
          Vicuna-16K& 25.0& \textbf{37.0}&36.0  &66.9  & 66.7 \\

         Vicuna+Retrieval&54.5 & 33.0 & \textbf{39.0} &  67.6&66.7   \\
         Vicuna+Truncation& \textbf{55.8} &  33.0& 32.0 & 66.7 &66.7  \\
         Vicuna+Summarization&43.5&28.0&25.0&\textbf{70.9}&\textbf{66.9}  \\
        \bottomrule
    \end{tabular}}
    \caption{Comparison of short-context LLMs with context compression techniques and long-context LLMs on selected tasks in BAMBOO-16k. We abbreviate AltQA, PaperQA, MeetingQA, Senhallu, and AbsHallu as AQA, PQA, MQA, SHallu, and AHallu, respectively.}
    \label{table_compression}
\end{table}

\begin{figure*}[htb]
    \centering
    
    \begin{minipage}[t]{0.48\linewidth}
        \includegraphics[width=\textwidth]{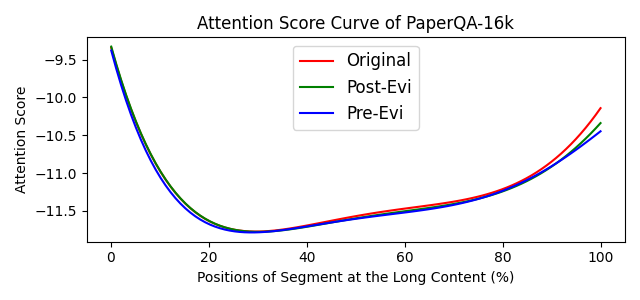}
    \end{minipage}
    \begin{minipage}[t]{0.48\linewidth}
        \includegraphics[width=\textwidth]{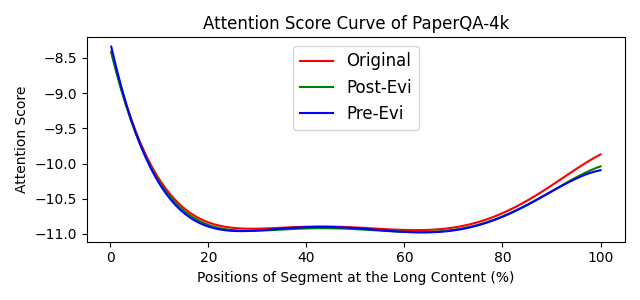} 
    \end{minipage}
    
    \caption{Logarithmic value of attention scores in three settings of evidence positions.}
    \label{fig_evidence}
\end{figure*}



The effect of different compression techniques is shown in Table~\ref{table_compression}.
Evidently, retrieval-augmented LLMs can achieve comparable or superior performance to long-context LLMs, aligning with the findings in \citet{xu-arxiv-2023-retrieval}. However, truncation and summarization methods often perform poorly due to the omission of substantial relevant information.



\subsubsection*{RQ5: Why ``Lost in The Middle'' Happens?}

Previous study~\cite{Liu-arxiv-2023-Lost} has observed that optimal performance is often attained when evidence information is positioned at the beginning or end of the input. To validate this phenomenon, we conducted experiments by placing evidence at the beginning or end of the long document (denoted as ``Pre-Evi'' and ``Post-Evi'', respectively). As illustrated in Table~\ref{table_lost}, despite disrupting the paper's original discourse structure, we consistently observed similar or improved performance when the evidence was situated at either end of the paper. Moreover, the performance gains were more substantial for longer input sequences.

To further investigate the rationale behind these remarkable properties, we analyze the attention maps. We compute the average attention scores for predictions across all heads and layers, and then partition the sequence into chunks of 32 tokens. Subsequently, we employ curve fitting to model the average scores of each chunk, as shown in Figure~\ref{fig_evidence}. Regardless of the location of the evidence within the input, we observe a U-shaped curve with higher attention allocated to tokens positioned at the two ends of the content. Consequently, it becomes evident that LLMs tend to utilize information located at the beginning or end of the input more effectively.

\begin{table}[htb]
    \centering
    \resizebox{\columnwidth}{!}{
    \begin{tabular}{ccccccc}
         \toprule
         \multirow{2.5}{*}{\textbf{Model}}   & \multicolumn{3}{c}{\textbf{PaperQA-16k}} & \multicolumn{3}{c}{\textbf{PaperQA-4k}}   \\ 
\cmidrule(r){2-4}\cmidrule(r){5-7} 
&Pre-Evi&Original&Post-Evi&Pre-Evi&Original&Post-Evi\\\midrule
         ChatGPT-16k&78.0&75.0&77.0&77.0&78.0&79.0\\
         Vicuna-16k&44.0&37.0&39.0&29.0&29.0&30.0
         \\\bottomrule
    \end{tabular}}
    \caption{Impact of evidence position.}
    \label{table_lost}
\end{table}

        
        
    




\subsection{Discussions}

Finally, we give discussions about the problems of LLMs over long texts. 

$\bullet$ \textbf{A severe problem is the cataclysmic forgetting of instructions.}  Frequently, the generating responses fail to align with the task demands. During the generation process, the instruction may be forgotten or even not understood at all, especially when the instructions are put at the beginning of long inputs. 
Smaller LLMs are more prone to adhering to the instructions, possibly owing to limitations in memorization. Thus, more robust long instruction datasets are required to enhance the instruction-following ability of long-context models


$\bullet$  \textbf{LLMs are prone to format errors in long text tasks.} In certain cases, the predictions of LLMs may be presented in an informal format, though they convey the same meaning as the answers. The models may offer additional explanations and repetitions or adopt different formats. Thus, they can not be identified by the evaluation code.
To alleviate the issue, we suggest reducing the format problems during the fine-tuning or RLHF stages and employing post-process techniques to rewrite the responses.

$\bullet$ \textbf{LLMs exhibit poor performance beyond the long texts.} Provided evidence or condensed texts, LLM may still generate false responses and make the same fault as inputting the complete long texts. This error can be attributed to insufficient ability, across various lengths and tasks. Thus, it is imperative to augment the overall capabilities of LLMs, rather than concentrating solely on the context of long text domains.


$\bullet$ \textbf{LLMs perform poorly on uncommon tasks.} Notably,  LLMs generally exhibit poor performance on tasks that are uncommon, \eg text sorting and code completion, and even underperform by random baselines. Most open-sourced LLMs were fine-tuned on multi-turns of conversation, question answering, and summarization tasks. The limited diversity in fine-tuning data hinders LLMs' generalization to uncommon long text tasks. We believe that broadening the variety of tasks and domains on fine-tuning data can better comprehensively enhance LLMs' long text modeling capacities.
\section{Conclusion}

We propose BAMBOO, a benchmark for comprehensively evaluating the long text modeling capabilities of LLMs. 
BAMBOO consists of five tasks with two length levels, enabling the evaluation of LLMs' main capacities across various dimensions and domains. 
Based on the evaluation of several long-context models on BAMBOO, we give an overall analysis of the performances of different models and tasks. Additionally, we analyze key questions of long text modeling, discuss the problems of long-context LLMs, and suggest directions for improving long text comprehension abilities.
We believe BAMBOO  has the potential to serve as a valuable tool for analyzing the comprehensive capacities and promoting the enhancement of long text modeling abilities of LLMs in the future.

\section{Acknowledgments}
This work was partially supported by National Natural Science Foundation of China under Grant No. 62222215, Beijing Natural Science Foundation under Grant No. 4222027 and L233008. Xin Zhao is the corresponding author.
\nocite{*}
\section{Bibliographical References}\label{sec:reference}

\bibliographystyle{lrec_natbib}
\bibliography{lrec-coling2024-example}
\section{Language Resource References}
\label{lr:ref}
\bibliographystylelanguageresource{lrec-coling2024-natbib}
\bibliographylanguageresource{languageresource}

\end{document}